\documentclass[10pt,twocolumn,letterpaper]{article}

\usepackage{iccv}
\usepackage{times}
\usepackage{epsfig}
\usepackage{graphicx}
\usepackage{amsmath}
\usepackage{amssymb}

\usepackage{booktabs}

\usepackage[pagebackref=true,breaklinks=true,letterpaper=true,colorlinks,bookmarks=false]{hyperref}

\iccvfinalcopy %

\def\ie{\emph{i.e.}}
\def\eg{\emph{e.g.}}
\def\etc{\emph{etc}}
\def\etal{{\em et al.~}}

\usepackage{multirow}

\usepackage{enumitem}
\usepackage{tabularx}
\usepackage{multirow}
\usepackage{stfloats}
\usepackage{threeparttable}
\usepackage[usenames,dvipsnames]{xcolor}

\usepackage{bm}

\def\ie{\emph{i.e.}}
\def\eg{\emph{e.g.}}
\def\etc{\emph{etc}}
\def\etal{{\em et al.~}}

\newlength\savedwidth

\definecolor{darkergreen}{RGB}{21, 152, 56}
\definecolor{red2}{RGB}{252, 54, 65}

\usepackage{xcolor}
\definecolor{yl_color}{RGB}{128, 255, 0}

\definecolor{blue2}{RGB}{20, 54, 254}

\usepackage{bbding}

\usepackage[american]{babel}
\usepackage{microtype}
\usepackage{cleveref}
\usepackage{adjustbox}

\usepackage{subcaption}

\usepackage{makecell}

\def\iccvPaperID{***} %

\ificcvfinal\fi

\begin{document}

\title{Looking and Listening: Audio Guided Text Recognition}

\author{
  Wenwen Yu$^{1}$,
  Mingyu Liu$^{1}$,
  Biao Yang$^{1}$,
  Enming Zhang$^{1}$,
  Deqiang Jiang$^{2}$,\\ 
  Xing Sun$^{2}$,
  Yuliang Liu$^{1}$, 
  Xiang Bai$^{1}$      \\
    $^1$Huazhong University of Science and Technology ~~~ $^2$Tencent YouTu Lab\\
    \small{\texttt{\{wenwenyu,ylliu\}@hust.edu.cn}} %
  }

\maketitle
\ificcvfinal\thispagestyle{empty}\fi

\begin{abstract}  
Text recognition in the wild is a long-standing problem in computer vision. Driven by end-to-end deep learning, recent studies suggest vision and language processing are effective for scene text recognition. Yet, solving edit errors such as add, delete, or replace is still the main challenge for existing approaches. 
In fact, the content of the text and its audio are naturally corresponding to each other, 
\textit{i.e.}, a single character error may result in a clear different pronunciation. 
In this paper, we propose the AudioOCR, a simple yet effective probabilistic audio decoder for mel spectrogram sequence prediction to guide the scene text recognition, which only participates in the training phase and brings no extra cost during the inference stage.
The underlying principle of AudioOCR can be easily applied to the existing approaches.
Experiments using 7 previous scene text recognition methods on 12 existing regular, irregular, and occluded benchmarks demonstrate our proposed method can bring consistent improvement. More importantly, through our experimentation, we show that AudioOCR possesses a generalizability that extends to more challenging scenarios, including recognizing non-English text, out-of-vocabulary words, and text with various accents.
Code will be available at https://github.com/wenwenyu/AudioOCR.
\end{abstract}

\section{Introduction}
Reading text from a natural image is a long-standing and profound problem in computer vision. It facilitates various applications such as document scanning, automatic assisting, and archiving data entry. Driven by the success of the deep neural network, text recognition has achieved remarkable progress in the past decades. 

Scene text recognition methods usually adopt visual processing and language modeling pipelines for parsing text content. The former usually adopts deep feature extraction for sequential prediction, while the latter is then used to rectify the recognition results both implicitly and explicitly. 
Although such a pipeline greatly advances the development of the field of scene text recognition, the edit errors such as add, delete, or replace are still one of the main challenges. 

\begin{figure}[tbp]
\centering
\includegraphics[width=0.48\textwidth]{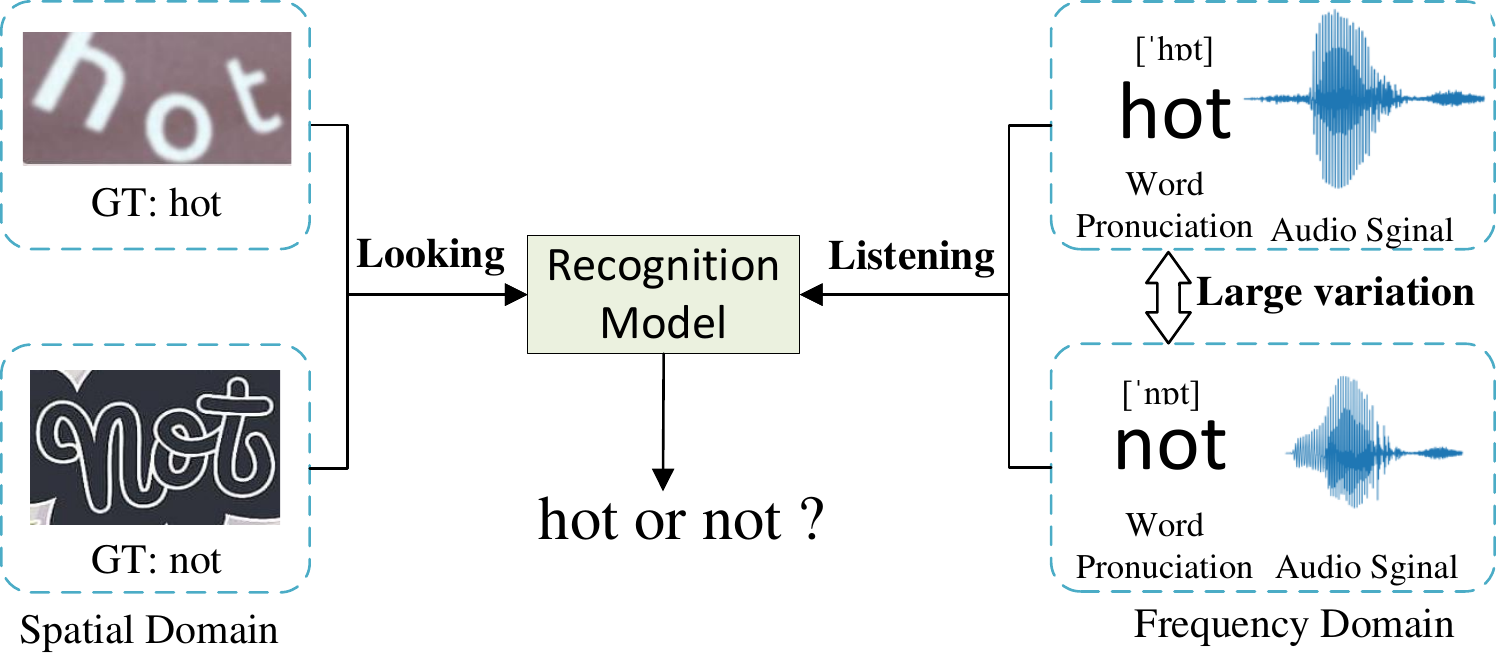}
\vspace{-1.5em}
\caption{Audio information can provide the salient information for guiding text recognition.}
\vspace{-1.8em}
\label{fig:intro_compare}
\end{figure}

To address this issue, we attempt to use audio information to guide the scene text recognition.
In fact, as shown in Figure~\ref{fig:intro_compare}, a single character such as ``h'' or ``n'' may not have a significant difference simply based on the visual appearance, as scene text could be presented with various fonts, vague, perspective distortion, \etc; however, such error may result in completely different of the pronunciation of the entire word. Recently, using audio information to assist vision tasks have been proven effective \cite{Vasudevan2022SoundAV,Braga2022BestOB,Cheng2020LookLA,Wu2020NotOL}, 
\eg, Tian \etal~\cite{Tian2021CanAI}. 
show that using audio-visual integration can strengthen the perception ability and thus improve the robustness of audio-visual models. Tang \etal~\cite{Tang2022YouCE} uses an Automatic Speech Recognition (ASR) model~\cite{Baevski2020wav2vec2A, Amodei2016DeepS2}
to convert annotations from audio modal to text modal for text spotting tasks, and accelerate the annotation procedure.
To the best of our knowledge, few methods consider using audio information for guiding scene text recognition.

In this paper, we propose a novel module, termed AudioOCR, using a transformer-based probabilistic audio decoder for spectrogram sequence prediction. It can be easily plugged into the existing approaches only during the training phase, and thus it brings no extra cost to the baseline model yet introduces obvious improvement. 
To enable the training, as shown in Figure~\ref{fig:mel_spec_label}, we adopt a PaddleSpeech~\cite{zhang2022paddlespeech}\footnote{\url{https://github.com/PaddlePaddle/PaddleSpeech}} engine to convert the text label to the target of mel spectrogram recognition. The text labels are directly from the existing synthesized datasets like MJ~\cite{Jaderberg2014SyntheticDA} and ST~\cite{gupta2016synthetic}, which are commonly served as standard training data for scene text recognition.
The underlying principle of our method is that the text always has the correct pronunciation regardless of its shape, size, clarity, \etc. We summarize our method as follows:

\begin{itemize}
\item We discovered a new approach, termed AudioOCR, whereby speech information can supplement text recognition, paving the way for further advances in the field.
\item The simplicity of our AudioOCR allows it to be applied to the existing approaches, \ie, it can apparently improve the performance of the recognizers, incurring acceptable computational overhead during the training phase and none whatsoever during the testing phase.
\item We evaluate our method by integrating the AudioOCR into 7 previous scene text recognition methods on 12 benchmarks including regular, irregular, and occluded datasets, which demonstrates our method can improve the performance of each recognition method by an average of 1.75\% in terms of the recognition accuracy.
\end{itemize}

\section{Related Works}
\label{subsec:rela}
\noindent\textbf{Visual-audio modeling.} In the visual-audio domain, there are several advancements that happened in visual-audio encoders in the past. Qu \etal\cite{Qu2021LipSound2SP} proposed an encoder-decoder architecture LipSound2 which directly predicts mel-scale spectrogram from raw face sequences. Chung \etal\cite{Chung2016LipRS} proposed a `Watch, Listen, Attend and Spell' (WLAS) network that learns to transcribe videos of mouth motion to characters for lip reading. Arandjelovic \etal\cite{Arandjelovi2017ObjectsTS} designed new network architectures that can be trained for cross-modal retrieval and localizing the sound source in an image through audio-visual correspondence tasks. Oh \etal\cite{Oh2019Speech2FaceLT} proposed Speech2Face model for reconstructing a facial image of a person from a short audio recording of that person speaking. 

\noindent\textbf{Text-audio modeling.} In the text-audio domain, it's a rapidly evolving field with ongoing research in many different areas, such as Automatic Speech Recognition (ASR)~\cite{Hannun2014DeepSS,Chorowski2015AttentionBasedMF,Baevski2020wav2vec2A,Li2021RecentAI} and Text-to-Speech (TTS) Synthesis~\cite{Wang2017TacotronTE,Tachibana2017EfficientlyTT,Ren2019FastSpeechFR,Ren2020FastSpeech2F}.

Compared to the visual-audio and text-audio domain, our method is different in several aspects: 1) The audio-decoder is a Transformer-based model that is parallelly trained via self-attention mechanism; 2) our method takes the input of visual modality and outputs the audio modality information; 3) our method develops a simple yet effective plug-and-play pipeline specifically designed for OCR, which does not introduce computational overheads in inference stage. 
 
\noindent\textbf{Scene text recognition.} 
Scene text recognition has been an active and challenging research topic in the field of computer vision. The popular methods have relied on encoder-decoder architectures \cite{ctc1,ar,energy,Baek2019WhatIW,Wang2020DecoupledAN,Bautista2022SceneTR,Wang2022MultiGranularityPF,Xie2022TowardUW,Bautista2022SceneTR}, and recent approaches have utilized convolutional neural networks (CNNs) as encoders, followed by recurrent neural networks (RNNs) or connectionist temporal classification (CTC) for decoding \cite{dtrn,crnn,ar}. Then attention mechanisms~\cite{r2am} and spatial transformer networks~\cite{rectify,aster} based methods have been proposed to handle irregular text with arbitrary shapes and rotations. Apart from this, some methods~\cite{masktextspotter1,textscanner} attempt to recognize text via character segmentation.

Recently, some methods tend to leverage auxiliary information to assist the recognition process~\cite{srn,Wang2021FromTT,abinet,seed,entirecnn,Tan2022PureTW,Na2021MultimodalTR}, such as the linguistic knowledge. SRN~\cite{srn} introduces a global semantic reasoning module to refine vision model predictions. SEED~\cite{seed} designs a semantic module, supervised by embedding outputs of a pretrained language model. Fang \etal\cite{entirecnn} propose a method that is based entirely on CNN for both vision and language models. ABINet~\cite{abinet} utilizes a bidirectional network as the language model and an explicit language model by blocking gradient flow. Besides, Song \etal\cite{Song2022VisionLanguagePF} mask random characters in the visual model and restore them by language model to learn linguistic knowledge in a union.

As shown above, previous works mainly leverage linguistic knowledge to assist recognition task; 
however, audio, another meaningful information is less studied in the literature. 
Thus, we explicitly utilize audio information through automatic generation without extra annotation costs.

\section{Methodology}
\label{sec:method}
In this section, we first introduce the overall architecture of the proposed audio guided text recognition framework. Then, we elaborate on the proposed plug-and-play audio decoder, termed AudioOCR, and further analyze the advantages of audio in feature modeling. Finally, we describe the details of the generation of the mel spectrogram of the audio decoder as well as the training process of cooperating with existing recognition methods.

\begin{figure}[htbp]
\centering
\includegraphics[width=0.49\textwidth]{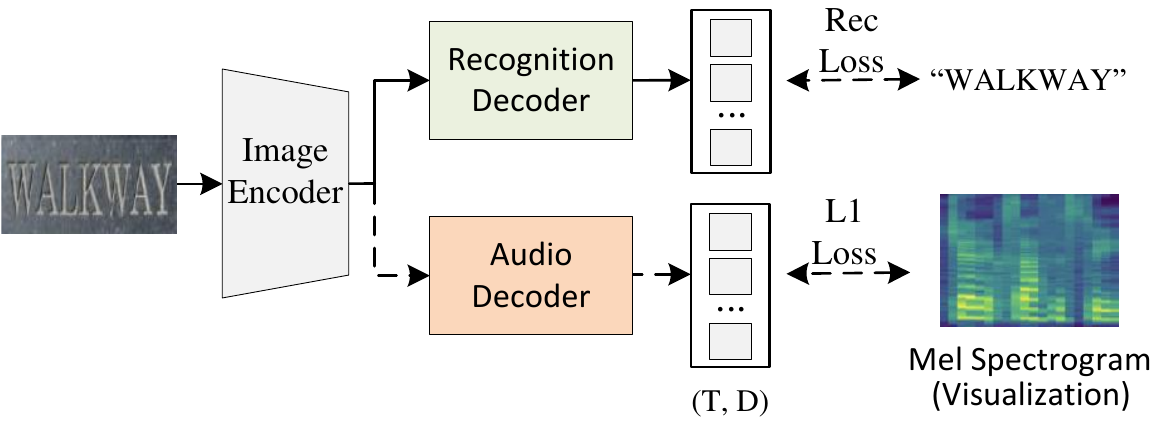}
\vspace{-1.5em}
\caption{The pipeline of audio guided text recognition method. AudioOCR consists of the image encoder, the recognition decoder, and the pluggable audio decoder for guiding text recognition. Dashed arrows are the training-only operators. Solid arrows indicate forward pass in both training and inference.}
\vspace{-1.5em}
\label{fig:method}
\end{figure}

\subsection{Audio Guided Text Recognition}
Our main objective is to precisely recognize each character in images with the help of not only looking but also listening. To this end, we jointly train a simple audio decoder with existing methods to predict the spectrogram of the audio of the corresponding transcription. As shown in Figure~\ref{fig:method}, our proposed audio-guided text recognition architecture is composed of an image encoder, a recognition decoder, and an audio decoder.

Given an input image $\mathbf{X}$, the image encoder generates visual image embeddings $\mathbf{\hat{X}}$, 
which is formulated as follows:
\begin{equation}
\mathbf{\hat{X}}
 = \mathtt{ImageEncoder}\left(\mathbf{X}~;  \Theta_\text{enc}\right)\,,
 \label{eq:image_encoder}
\end{equation}
where $\mathbf{X} \in \mathbb{R}^{ H' \times W' \times 3 } $ denotes the vector of input image, $H'$ and $W'$ represent the height and width of image, respectively. $\mathbf{\hat{X}} \in \mathbb{R}^{  H \times W \times d_{model}}$ represents the output of the image encoder, and $d_{model}$ indicates the dimension of model. $\mathbf{\hat{X}}$ will be used as input for the recognition decoder and audio decoder. $\Theta_\text{enc}$ represents the parameters of the image encoder. In practice, image encoder, depending on different recognizers, can be instantiated into a wide variety of model structures, 
which usually includes the CNN based \eg~ResNet~\cite{He2016DeepRL} and transformer~\cite{Vaswani2017AttentionIA} based image encoder along with the extra module \eg, flexible rectification~\cite{aster}.

The recognition decoder receives the visual image embedding $\mathbf{\hat{X}}$
to produce a sequence of predicted characters $\hat{Y}=\{y_1,y_2,\ldots,y_n\}$, in which $n$ is the length of predicted sequence and $y_i$ is the $i$-th predicted characters, which can be expressed as:
\begin{equation}
\hat{Y}
 = \mathtt{RecoDecoder}\left(\mathbf{\hat{X}}~;  \Theta_\text{rec}\right)\,,
 \label{eq:reco_encoder}
\end{equation}
where RecoDecoder means the recognition decoder module, and $\Theta_\text{rec}$ represents the parameters of the recognition decoder. 

In practice, there are three types of recognition decoders commonly used in real scenarios, including (1) Connectionist Temporal Classification (CTC)~\cite{ctc1,crnn}; (2) attention-based sequence prediction (Attn) \cite{aster,Li2019ShowAA}; and (3) transformer decoder based sequence prediction (Trans) \cite{Lee2020OnRT,Lu2021MASTERMN,peng2022spts}. CTC allows for the prediction of a non-fixed number of a sequence even though a fixed number of the features are given. Attn-based decoder automatically captures the information flow within the input sequence to predict the output sequence. It enables a text recognition model to learn a character-level language model representing output class dependencies. Trans-based decoder leverages self-attention to model long-range dependency between features and predicts the characters of the sequence via an auto-regressive manner.

\subsection{Audio Decoder}
\begin{figure}[htbp]
\centering
\includegraphics[width=0.49\textwidth]{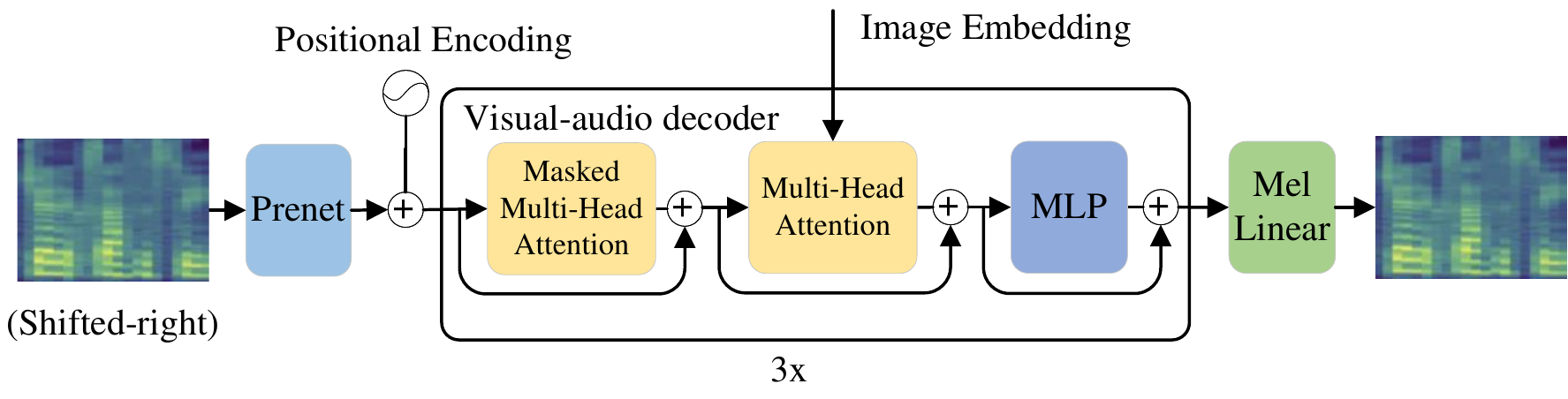}
\vspace{-1.5em}
\caption{The architecture of the audio decoder. It includes a Prenet, a visual-audio decoder, and a Mel Linear. LayerNorm and Dropout layers are omitted due to space constraints.}
\vspace{-1.5em}
\label{fig:audio_decoder}
\end{figure}

The audio decoder, as shown in Figure~\ref{fig:audio_decoder},
is proposed to guide the recognizers to capture audio modal information. It consists of a Prenet, a visual-audio decoder, and a Mel Linear.

Prenet is composed of two fully connected layers (each has 256 hidden units) with ReLU activation. Prenet is responsible for projecting mel spectrograms into the phoneme subspace of audio, and thus the similarity of a phoneme and mel frame pair can be measured to force the following visual-audio decoder to easily capture visual and audio interaction, which is defined as:
\begin{equation}
\mathbf{\hat{Z}}
 = \mathtt{Prenet}\left(\mathbf{\dot{Z}}~;  \Theta_\text{pre}\right)\,,
 \label{eq:prenet}
\end{equation}
where $\mathbf{\dot{Z}} $ is the shifted-right format~\cite{Vaswani2017AttentionIA} of the mel spectrogram $\mathbf{Z} \in \mathbb{R}^{ T \times 80  } $ (Sec.~\ref{sec:spec_label_generation} describes the generation way), where $T$ is the time length of audio. $\mathbf{\hat{Z}} \in \mathbb{R}^{ T \times 256  } $ denotes the transformed vector of mel spectrogram containing compact and low dimensional subspace features of audio. $\Theta_\text{pre}$ represents the parameters of the Prenet.

The visual-audio decoder is implemented by the transformer decoder. It utilizes the mel spectrogram features $\mathbf{\hat{Z}}$ as query and image embeddings $\mathbf{\hat{X}}$ as key and value to model cross-modal interaction between visual and audio features for learning effective scene text visual representations via cross-attention. The visual-audio decoder hence learns the relationships between visual and audio features. To this end, the visual features $\mathbf{\hat{X}}$ implicitly contain the audio information which is helpful for text recognition of complex scenarios. We add a set of learned positional encoding~\cite{Vaswani2017AttentionIA} to mel spectrogram $\mathbf{\hat{Z}}$ for capturing the sequential information, then passing it to visual-audio decoder to refine the mel spectrogram conditioned on visual features:
\begin{equation}
\mathbf{\Tilde{Z}}
 = \mathtt{VADecoder}\left(\mathbf{\mathtt{Q}=\hat{Z}, \mathtt{K}=\hat{X}, \mathtt{V}=\hat{X}}~;  \Theta_\text{va}\right)\,,
 \label{eq:visual_audio_decoder}
\end{equation}
where $\mathbf{\Tilde{Z}} \in \mathbb{R}^{ T \times 256  } $ denotes the refined mel spectrogram. VADecoder means the visual-audio decoder. $\Theta_\text{va}$ represents the parameters of the visual-audio decoder. In practice, the default settings of the visual-audio decoder consist of $N=3$ stacked decoder layers. Each decoder layer contains a masked multi-head attention, a multi-head attention layer, and an MLP network.

The Mel Linear implemented by a linear projection and the $\mathbf{\Tilde{Z}}$ is then used to predict the mel spectrogram, which can be formulated by:
\begin{equation}
\mathbf{\Bar{Z}}
 = \mathtt{MelLinear}\left(\mathbf{\Tilde{Z}}~;  \Theta_\text{mel}\right)\,,
 \label{eq:mel_linear}
\end{equation}
where $\mathbf{\Bar{Z}} \in \mathbb{R}^{ T \times 80  } $ denotes the predicted mel spectrogram. MelLinear means the Mel Linear. $\Theta_\text{mel}$ represents the parameters of the Mel Linear. Then $\mathbf{\Bar{Z}} $ and $\mathbf{Z}$ will calculate the \textit{L1} loss $L_{mel}$:
\begin{equation}
    L_{mel}
 = \mathtt{L1}\left(\mathbf{\check{Z}}~, \mathbf{\Bar{Z}}\right)\,,
    \label{eq:L_mel}
\end{equation}
where the mel spectrogram label $\mathbf{\check{Z}}$ and the input of Prenet $\mathbf{\dot{Z}}$ are basically the same (mel spectrogram $\mathbf{Z}$). The only difference is the input $\mathbf{\dot{Z}}$ is the shifted-right format~\cite{Vaswani2017AttentionIA} of the $\mathbf{Z}$.

\begin{figure}[htbp]
\centering
\includegraphics[width=0.49\textwidth]{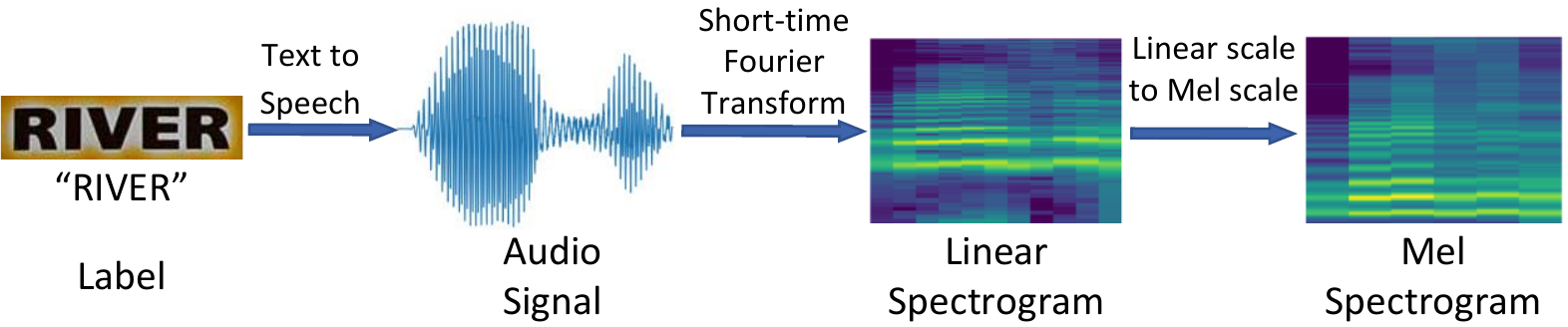}
\vspace{-1.5em}
\caption{The pipeline of converting the text label to the mel spectrogram used in the audio decoder. }
\vspace{-1.5em}
\label{fig:mel_spec_label}
\end{figure}

\subsection{Mel Spectrogram $\mathbf{Z}$ Generation}\label{sec:spec_label_generation}
Since there is no publicly available dataset that includes text-audio pairs, we use the off-the-shelf text-to-speech tool PaddleSpeech to generate the audio of the transcription of every image and store the audio file with the suffix $.wav$. We then convert the audio data to the mel spectrogram used for the training of the audio decoder.

Specifically, we first get the waveform $\bm{x}$ of the audio from the $.wav$ file, with the shape of $[C, T]$ where $C$ is the number of channels ($C = 2$ for stereo audio), and $T$ the number of points used to represent the audio, as shown in Figure~\ref{fig:mel_spec_label}. Then Short-Time Fourier Transform (STFT) is employed over the waveform $\bm{x}$ to represent audio into a complex-valued tensor of shape $[C, F, L]$ where $F$ is the number of frequencies and $L$ is the length. The STFT is computed on overlapping windowed segments of the waveform $\bm{x}$. The STFT at a point ($f$, $l$) is defined as: 
\begin{equation}
\mathtt{STFT}\{\bm{x}\}(f, l)=\sum_{t=0}^{T}\bm{x}_t\cdot \bm{w}_{t-l}\cdot \mathtt{exp}\left(-i \frac{2 \pi t}{T} f \right) \,,
 \label{eq:stft}
\end{equation}
where $\bm{w}$ is a slide window function of length $W$ that is used to isolate the chunk of interest from the waveform signal. $f$, $l$, and $i$ are frequency, time step, and imaginary part, respectively.

Instead of real and imaginary parts, often the STFT is converted
into magnitude and phase, a representation that can be inverted back to real and imaginary parts, and hence back to the waveform. Magnitude, called \textit{linear sepctrogram}, is defined as $|\mathtt{STFT}\{\bm{x}\}(f, l)|$.

Finally, we perform a mathematical operation on the frequencies of the linear spectrogram to convert the linear scale to the mel scale~\cite{Stevens1937ASF}, called \textit{mel spectrogram}. The reason is that humans do not perceive frequencies on a linear scale~\cite{Stevens1937ASF}, and equal distances in mel scale sound equally distant to the listener. The mathematical operation is formulated as:
\begin{equation}
m=2595 * \log _{10}\left(1+\frac{f}{700}\right)\,,
 \label{eq:mel_convert}
\end{equation}
where $f$ denotes the linear scale frequency, and $m$ denotes the mel scale frequency. The configuration of mel spectrogram generation refers to appendix. %

\subsection{Training strategy}
Our model is designed to jointly learn with recognition loss and audio loss.
The loss function $L$ can be expressed as a weighted sum of the recognition losses and audio losses:
\begin{equation}
    L=L_{rec}+\alpha \times L_{mel},
    \label{eq:loss}
\end{equation}
where $L_{rec}$ is the loss for the recognition decoder, and $L_{mel}$ (Eq.~\ref{eq:L_mel}) is the \textit{L1} loss for the audio decoder. According to the numeric values of the losses, $\alpha$ is set to $1.0$ by default.

\section{Experiment}
\label{sec:exp}
We first incorporate the AudioOCR into the existing recognizers to validate its effectiveness. Next, we conduct ablation studies to measure the sensitivity of the designs and generalization ability for AudioOCR. 

\subsection{Datasets}
We use three types of datasets including synthetic datasets, real scene text recognition benchmarks, and our SynthAudio data. Synthetic datasets including MJSynth (MJ)~\cite{Jaderberg2014SyntheticDA}, SynthText (ST)~\cite{gupta2016synthetic}, and SynAdd (SA)~\cite{Li2019ShowAA}. The details descriptions of the synthetic datasets refer to appendix.

\textbf{Scene text recognition benchmarks} including
(1) IIIT5k-Words (IIIT)~\cite{mishra2012top}; (2) Street View Text (SVT)~\cite{Shi2014EndtoendST} ; (3) ICDAR 2013 (IC13) ~\cite{Karatzas2013ICDAR2R}; (4) ICDAR 2015 Incidental Text (IC15) ~\cite{Karatzas2015ICDAR2C}; (5)
SVTP-Perspective (SVTP)~\cite{Phan2013RecognizingTW}; (6) CUTE80 (CT80)~\cite{Risnumawan2014ARA}; (7) COCOText-Validation (COCO)~\cite{veit2016coco}; (8) CTW dataset~\cite{Liu2019CurvedST}; (9) Total-Text dataset (TT)~\cite{Chng2017TotalTextAC}; (10) Occlusion Scene Text (OST)~\cite{Wang2021FromTT}, including
weakly occluded scene text (WOST) and heavily occluded scene text (HOST) are used to evaluate our recognition models; and (11) WordArt dataset~\cite{Xie2022TowardUW}. The detailed descriptions of the benchmarks refer to appendix. %

\textbf{SynthAudio data} is generated using the open-source text-to-speech toolkit PaddleSpeech~\cite{zhang2022paddlespeech}. Our corpus is directly from the label of public datasets, including synthetic datasets ST, MJ, real datasets COCO-Text, IIIT5K, SVT, \etc. The detailed information of SynthAudio refers to appendix. %

\subsection{Implementation Details}
Our model is implemented based on MMOCR~\cite{mmocr2021}\footnote{\url{https://github.com/open-mmlab/mmocr}}. 
The model is trained using 4 NVIDIA GTX 3090 GPUs.
We choose to perform experiments on the existing reproduced models on MMOCR, including MASTER~\cite{Lu2021MASTERMN}, CRNN~\cite{crnn}, SAR~\cite{Li2019ShowAA}, SATRN~\cite{Lee2020OnRT}, Robust\_Scanner (Robust\_S)~\cite{yue2020robustscanner}, ABINet~\cite{abinet}, and ASTER~\cite{abinet}, which contains CTC, Attention and Transformer based recognizers.

For the training dataset, we keep the same configuration as the original paper. For different models, we have different data configurations, as shown in Table~\ref{tab:cooperation_with_recognizers}. Unless specified, we set the number of audio decoder layers as 3. We report the word accuracy metric (\%) in all experiments.

\subsection{Cooperation with Existing Recognizers} As illustrated in Table~\ref{tab:cooperation_with_recognizers}, we validate the benefits of the pluggable AudioOCR for existing recognizers. 
Specifically, we evaluate AudioOCR by combining it with 7 scene text recognition methods on 12 benchmarks including regular, irregular, and occluded datasets.
Equipping with audio decoder methods achieves consistent improvement compared to baseline methods in average recognition accuracy. It is worth mentioning that such improvement is free in the inference stage without introducing any extra costs, while for the training stage, it only brings an additional 4.6 ms/image on average for computation overhead as shown in Table~\ref{tab:training_time}.

\begin{table*}[htbp!]
\centering
\small
\setlength\tabcolsep{1.7pt}
{
\begin{tabularx}{1.0\linewidth}{c|cccccccccccccccc}
\hline  
 \multicolumn{1}{c}{\multirow{1}{*}{}} & \multirow{2}{*}{Method} & \multirow{2}{*}{Img. data} & \multirow{2}{*}{Audio data} & \multicolumn{3}{c}{Regular} & \multicolumn{6}{c}{Irregular}        & \multicolumn{2}{c}{Occluded} & \multirow{2}{*}{WordArt} & \multirow{2}{*}{Avg.} \\
 \cmidrule(lr){5-7} \cmidrule(lr){8-13} \cmidrule(lr){14-15} 
 \multicolumn{1}{c}{\multirow{1}{*}{}}  &    &    &   & IIIT5K    & SVT    & IC13   & IC15 & SVTP & CT80 & CTW & TT & COCO & HOST          & WOST         &                          \\ \hline
\multirow{3}*{\rotatebox{90}{CTC}} & CRNN-paper              & MJ                        & -                           & 78.2     & 80.8    & 86.7   & -     & -     & -     & -     & -     & -     & -             & -            & -                        & -                 \\
 & CRNN*          & MJ                        & -                           & 80.5     & \textbf{81.0}    & 86.5   & 54.1  & 59.1  & 55.6  & 54.3 & 52.2  & 37.1 & 30.1         & 45.7        & 33.1                    & 55.8          \\
 & CRNN                    & MJ                        & 600k                         & \textbf{82.0}    & \textbf{81.0}   & \textbf{88.3}  & \textbf{57.5} & \textbf{65.4} & \textbf{60.4} & \textbf{55.4} & \textbf{53.5} & \textbf{37.7} & \textbf{36.9}         & \textbf{53.0}        & \textbf{38.4}                    & \textbf{59.1}    \\ \hline
\multirow{9}*{\rotatebox{90}{Attention}} & ASTER-paper          & ST+MJ                     & -                           & 91.9    & 88.8   & 89.8  & -     & 74.1 & 73.3 &    -   &    -   &    -   &      -   &    -          &         -                 & -               \\
 & ASTER*           & ST+MJ                     & -                           & 92.5     & 89.0    & 91.2     & 73.3    & 80.2  & 82.3  &   69.0    &   69.6    &   52.3    &       40.4        &       61.4       &        58.2                  & 71.6               \\
 & ASTER                & ST+MJ                     & 4m                        & \textbf{93.7}    & \textbf{89.5}   & \textbf{93.2}   & \textbf{74.7} & \textbf{81.1} & \textbf{86.1} & \textbf{69.8} & \textbf{71.6}  & \textbf{53.1} & \textbf{47.8}          & \textbf{65.5}        & \textbf{60.6}                    & \textbf{73.9}               \\ \cline{2-17}
 & SAR-paper                 & S+R                & -                           & 95.0       & 91.2    & 94.0     & 78.8  & \textbf{86.4}  & 89.6  &   -    &    -   &   -    &    -           &     -         &    -                      & -          \\
 & SAR*             & S+R               & -                           & 95.0       & 89.6    & \textbf{93.7}   & 79    & 82.2  & 88.9  &    75.6   &  75.8     &    63.2   &        \textbf{41.4}       &   65.3           &                 63.3         & 76.1          \\
 & SAR                       & S+R              & 4m                        & \textbf{95.5}     & \textbf{90.9}    & 93.5   & \textbf{79.3}  & 82.6 & \textbf{89.9} &     \textbf{76.0}  & \textbf{78.9}      & \textbf{64.3}      &    41.2           &  \textbf{65.4}            &      \textbf{63.4}                    & \textbf{76.7}          \\ \cline{2-17}
 & Robust\_S-paper     & S+R            & -                           & \textbf{95.4}     & 89.3    & \textbf{94.1}   & 79.2  & \textbf{82.9}  & 89.6  &    -   &    -   &     -  &     -          &      -        &       -                   & -          \\
 & Robust\_S* & S+R                & -                           & 95.1     & 89.2    & 93.1   & 77.8  & 80.3  & 90.3  &   75.4    &   75.7    &    62.0   & 38.6              &      65.0        &         61.0                 & 75.3         \\
 & Robust\_S           & S+R                & 2m                        & 95.1     & \textbf{90.6}    & \textbf{94.1}   & \textbf{79.6}  & 81.4  & \textbf{91}    &   \textbf{75.5}    &   \textbf{77.9}    &  \textbf{\underline{64.6}}     & \textbf{42.4}              &       \textbf{66.1}       &              \textbf{62.4}            & \textbf{76.7}          \\ \hline
\multirow{9}*{\rotatebox{90}{Transformer}} & SATRN-paper               & ST+MJ                     & -                           & 92.8     & 91.3    & 94.1   & 79    & 86.5  & 87.8 &    -   &    -   &     -  &     -          &      -        &       -                   & -          \\
 & SATRN*                     & ST+MJ                     & -                        & 94.7    & 90.7   & 95.4  & 81.9 & 85.7 & 86.5 &    75.6   &    78.4   &    59.8  &     56.7        &      72.0        &       63.7                    & 78.4          \\
 & SATRN                     & ST+MJ                     & 4m                        & \textbf{95.4}    & \textbf{92.9}   & \textbf{96.6}  & \textbf{84.2} & \textbf{88.8} & \textbf{89.9} &    \textbf{76.2}       &     \textbf{79.1}  &     \textbf{59.9}          &      \textbf{64.7}        &      \textbf{76.1}       &    \textbf{67.5}         & \textbf{81.0}          \\ \cline{2-17}
 & MASTER-paper            & ST+MJ+SA              & -                           & 95       & 90.6    & 95.3   & 79.4  & 84.5  & 87.5  & -     & -     & -     & -             & -            & -                        & -          \\
 & MASTER*            & ST+MJ+SA              & -                           & 95.3    & 89.9    & 95.2  & 77.0 & 83.0 & 89.9 & 74.2 & 77.4 & 54.0 & 52.2         & 72.7        & 65.3                    & 77.2              \\
 & MASTER                  & ST+MJ+SA              & 4m                        & \textbf{95.7}    & \textbf{91.2}   & \textbf{95.9}  & \textbf{79.7} & \textbf{84.7} & \textbf{89.9} & \textbf{75.7}  & \textbf{80.2} & \textbf{55.5}  & \textbf{52.8}         & \textbf{74.2}        & \textbf{66.3}                    & \textbf{78.5}          \\ \cline{2-17}
 & ABINet-LV-paper              & ST+MJ                     & -                           & \textbf{\underline{96.2}}     & 93.5    & 97.4   & 86    & 89.3  & 89.2  &    -   &    -   &   -    &      -         &    -          &           -               & -          \\
 & ABINet-LV*          & ST+MJ                     & -                           & 95.7     & 94.6    & 95.7   & 85.1  & 89.3  & 90.3  &  76.7     &   81.2    &   63.7    &       62.8        &      76.5        &                  \textbf{\underline{66.7}}        & 81.5          \\
 & ABINet-LV                    & ST+MJ                     & 4m                        & 95.9     & \textbf{\underline{94.9}}    & \textbf{\underline{97.8}}   & \textbf{\underline{86.1}}  & \textbf{\underline{89.9}}  & \textbf{\underline{91.4}} &    \textbf{\underline{76.8}}   &    \textbf{\underline{82.8}}   &    \textbf{64.0}   &  \textbf{\underline{63.5}}      & \textbf{\underline{76.8}}      &           \textbf{\underline{66.7}}          & \textbf{\underline{82.3}}               \\ \hline

\end{tabularx}}
\vspace{-1em}
\caption{The results of cooperating with existing recognizers. * represents our reproduced results. S+R means ST+MJ+SA+real data. k: one thousand, m: one million. For ABINet, we use the same version of IC13 and IC15 as its paper. In each column, the best performance result of every method is shown in bold font, and the global best result is shown with an underline.}
\label{tab:cooperation_with_recognizers}
\vspace{-1.5em}
\end{table*}

\begin{table*}[htbp]
\centering
\small
\setlength\tabcolsep{2.6pt}
{
\begin{tabularx}{1.0\linewidth}{ccccccccccccccccc}
\hline
\multirow{2}{*}{Method} & \multirow{2}{*}{Layers} & \multirow{2}{*}{Img. data} & \multirow{2}{*}{Audio data} & \multicolumn{3}{c}{Regular} & \multicolumn{6}{c}{Irregular}                       & \multicolumn{2}{c}{Occluded} & \multirow{2}{*}{WordArt} & \multirow{2}{*}{Avg.} \\ \cmidrule(lr){5-7} \cmidrule(lr){8-13} \cmidrule(lr){14-15} 
                        &                                          &                           &                             & IIIT5K  & SVT     & IC13    & IC15   & SVTP   & CT80   & CTW    & TT     & COCO   & HOST          & WOST         &                          &                       \\ \hline
CRNN                    & 1                                        & MJ                        & 600k                         & 81.3    & 79.3   & 86.8   & \textbf{57.6}  & 62.9  & \textbf{60.4}  & \textbf{55.5}  & 51.0    & 37.3  & 33.7          & 48.1         & 37.0                       & 57.6          \\
CRNN                    & 2                                        & MJ                        & 600k                         & 81.0    & 80.2   & 87.8  & 56.7 & 62.5 & 60.1 & 54.7 & 52.1 & 38.0 & 33.3         & 48.6         & 36.4                     & 57.6       \\
CRNN                    & 3                                        & MJ                        & 600k                         & \textbf{82.0}    & \textbf{81.0}   & \textbf{88.3}  & 57.5 & \textbf{65.4} & \textbf{60.4} & 55.4 & \textbf{53.5} & 37.7 & \textbf{36.9}         & \textbf{53.0}        & \textbf{38.4}                    & \textbf{59.1}              \\
CRNN                    & 4                                        & MJ                        & 600k                         & 81.8    & 80.5   & 86.7   & 57.3 & 62.6 & \textbf{60.4} & 53.8 & 52.1 & \textbf{38.2} & 34.5         & 47.9        & 37.3                    & 57.8          \\ \hline
ASTER                   & 1                                        & ST+MJ                     & 600k                         & 93.3    & 88.7   & \textbf{93.0}     & 73.9  & \textbf{80.0}   & 83.7 & 69.1 & 70.3 & 52.7 & 46.34         & 65.0        & 59.6                    & 73.0              \\
ASTER                   & 2                                        & ST+MJ                     & 600k                         & 93.4    & 87.3   & 92.0  & 73.5 & 79.1 & 84.7 & 68.8 & 70.0 & 52.3 & 47.0         & 65.2        & 60.2                    & 72.8              \\
ASTER                   & 3                                        & ST+MJ                     & 600k                         & \textbf{93.6}     & \textbf{89.0}   & \textbf{93.0}     & \textbf{74.2} & \textbf{80.0}    & \textbf{85.4} & \textbf{69.7} & \textbf{71.0} & \textbf{53.1} & \textbf{47.1}          & \textbf{65.4}        & \textbf{60.2}                    & \textbf{73.5}          \\
ASTER                   & 4                                        & ST+MJ                     & 600k                         & 93.5     & 88.6   & \textbf{93.0}     & 73.4 & 79.8 & 84.0 & 69.7 & 70.7 & 52.2 & 47.0         & 65.3        & 60.1                    & 73.1 \\
\hline
ABINet                   & 1                                        & ST+MJ                     & 4m                         & 95.8    & 94.7   & 97.3     & 85.8  & 89.6   & 90.3 & \textbf{76.9} & 81.3 & 63 & 62.3         & 76.3        &  \textbf{66.7}                    & 81.6              \\
ABINet                   & 2                                        & ST+MJ                     & 4m                         & \textbf{95.9}    & 94.6   & \textbf{97.8}  & 85.9 & 89.5 & 90.3 & 76.7 & 81.2 & 63.7 & 62.5       & 76.4       & 65.9                  & 81.7             \\
ABINet                   & 3                                        & ST+MJ                     & 4m                         & \textbf{95.9}     & \textbf{94.9}   & \textbf{97.8}     & \textbf{86.1} & \textbf{89.9}    & \textbf{91.4} & 76.8 & \textbf{82.8} & 64 & \textbf{63.5}          & \textbf{76.8}        &  \textbf{66.7}                    & \textbf{82.2}          \\
ABINet                   & 4                                        & ST+MJ                     & 4m                         & 95.4     & 94.6   & 97.1     & 85.7 & 89.6 & 91 &      76.3 & 81.1 & \textbf{64.7} & 63.4         & 76.4        & 65.7                    & 81.75 \\
\hline
\end{tabularx}}
\vspace{-1em}
\caption{Ablation study of the depth of the audio decoder layer. 
Img. and Avg. are short for image and average, respectively. In each column, the best performance result of every method is shown in bold font.}
\label{tab:abl_layer}
\vspace{-1em}
\end{table*}

\begin{table*}[t]
\centering
\small
\setlength\tabcolsep{1.3pt}
{
\begin{tabularx}{1.0\linewidth}{ccccccccccccccccc}
\hline
\multirow{2}{*}{Method} & \multirow{2}{*}{Speaker} & \multirow{2}{*}{Img. data} & \multirow{2}{*}{Audio data} & \multicolumn{3}{c}{Regular} & \multicolumn{6}{c}{Irregular}        & \multicolumn{2}{c}{Occluded} & \multirow{2}{*}{WordArt} & \multirow{2}{*}{Avg.} \\ \cmidrule(lr){5-7} \cmidrule(lr){8-13} \cmidrule(lr){14-15} 
                        &                            &                           &                             & IIIT5K    & SVT    & IC13   & IC15 & SVTP & CT80 & CTW & TT & COCO & HOST          & WOST         &                          \\ \hline
CRNN                    & Female, British          & MJ                        & 600k             & 82.0    & \textbf{81.0}   & 88.3  &  \textbf{57.5} & 65.4 & 60.4 & \textbf{55.4} & 53.5 & 37.7 & \textbf{36.9}         & \textbf{53.0}        & \textbf{38.4}                    & \textbf{59.1}                           \\
CRNN                    & Male, British           & MJ                        & 600k                         & 81.8     & 79.9   & 88.4  & 57.0 & 62.6 & \textbf{63.1} & 54.9 & 52.3 & 37.3 & 33.9         & 47.6         & 36.7                    & 58.0          \\
CRNN                    & Male, American         & MJ                        & 600k                         & 81.4     & 79.3    & 87.0   & 55.6  & 64.0  & 58.6  & 55.3  & 52.4  & 37.5  & 33.2          & 48.8         & 36.6                     & 57.5          \\
CRNN                    & Female, American        & MJ                        & 600k                         &      \textbf{82.2}    &       80.4  &     \textbf{88.7}   &    57.1   &    \textbf{65.5}   &   62.5    &    54.7   &   \textbf{53.6}    &    \textbf{37.8}   &        36.4       &       52.6       &                 38.1         &       \textbf{59.1}                          \\ \hline
ASTER                   & Female, British         & ST+MJ                     & 600k                         & 93.6     & \textbf{89.0}   & \textbf{93.0}     & \textbf{74.2} & \textbf{80.0}    & \textbf{85.4} & \textbf{69.7} & \textbf{71.0} & \textbf{53.1} & \textbf{47.1}          & \textbf{65.4}        & \textbf{60.2}                    & \textbf{73.5}          \\
ASTER                   & Male, British           & ST+MJ                     & 600k                         & 93.5     & 87.1   & 92.3  & 73.9  & 77.5 & 85.3 & 69.2 & 70.3 & 52.8 & \textbf{47.1}         & 64.6        & 60.1                    & 72.8          \\
ASTER                   & Male, American         & ST+MJ                     & 600k                         & \textbf{93.7}    & 87.3   & \textbf{93.0}     & 73.1 & 79.2 & 85.0 & 69.2 & 69.6 & 52.3 & 46.4          & 64.6        & 60.1                    & 72.8          \\
ASTER                   & Female, American       & ST+MJ                     & 600k                         & \textbf{93.7}    & 88.1    & 92.7  & 73.2 & 79.3  & 84.0 & 69.3 & 70.6  & 52.3 & 47.0            & \textbf{65.4}        & \textbf{60.2}                    & 73.0    \\ \hline

ABINet                   & Female, British         & ST+MJ    &4m &  \textbf{{95.9}}&       \textbf{{94.9}}    & \textbf{{97.8}}   & \textbf{{86.1}}  & \textbf{{89.9}}  & \textbf{{91.4}} &    76.8   &    82.8   &    64.0   &  \textbf{{63.5}}      & \textbf{{76.8}}      &           66.7         & \textbf{{82.3}}          \\
ABINet                   & Male, British           & ST+MJ                     & 4m                         & 95.7&94.7&95.6&84.4&89.6&90.0&77.9&84.2&\textbf{65.2}&61.6&76.2&66.4&81.8          \\
ABINet                   & Male, American         & ST+MJ                     & 4m                         & 95.7&94.6&95.9&84.8&89.5&90.6&77.8&84.3&64.9&61.6&76.4&\textbf{66.8}&82.0          \\
ABINet                   & Female, American       & ST+MJ                     & 4m                         & 95.6&94.2&95.7&84.9&\textbf{89.9}&90.9&\textbf{78.1}&\textbf{84.5}&64.8&61.5&76&65.5&81.8   \\ \hline

\end{tabularx}}
\vspace{-1em}
\caption{Ablation study of using 4 different accents for the audio information. 
Img. and Avg. are short for image and average, respectively. In each column, the best performance result of every method is shown in bold font.}
\label{tab:abl_type}
\vspace{-1.5em}
\end{table*}

\begin{table*}[htbp]
\centering
\small
\setlength\tabcolsep{3.2pt}
{
\begin{tabularx}{0.98\linewidth}{cccccccccccccccc}
\hline
\multirow{2}{*}{Method} & \multirow{2}{*}{Img. data} & \multirow{2}{*}{Audio data} & \multicolumn{3}{c}{Regular} & \multicolumn{6}{c}{Irregular}        & \multicolumn{2}{c}{Occluded} & \multirow{2}{*}{WordArt} & \multirow{2}{*}{Avg.} \\ \cmidrule(lr){4-6} \cmidrule(lr){7-12} \cmidrule(lr){13-14} 
                        &                           &                             & IIIT5K    & SVT    & IC13   & IC15 & SVTP & CT80 & CTW & TT & COCO & HOST          & WOST         &                          \\ \hline
CRNN                    & MJ                        & 100k                         & 81.0    & 79.8    & 87.4  & 55.3 & 62.1 & 57.6 & 54.6 & 52.1 & 37.2 & 31.1         & 46.8        & 35.2                    & 56.7          \\
CRNN                    & MJ                        & 600k                         & \textbf{82.0}    & 81.0   & \textbf{88.3}  & \textbf{57.5} & \textbf{65.4} & 60.4 & \textbf{55.4} & \textbf{53.5} & 37.7 & \textbf{36.9}    & \textbf{53.0} & 38.4                    & 59.1              \\
CRNN                    & MJ                        & 4m                        & 81.9    & \textbf{81.2}   & 88.2  & 57.4 & 64.5  & \textbf{62.2} & 55.3 & 53.4 & \textbf{37.8} & 36.7         & 52.2        & \textbf{39.5} & \textbf{59.2}          \\
\hline
ASTER                   & ST+MJ                     & 100k                         & 93.2    & 88.2   & 93.1 & 73.9 & 79.3 & 82.9 & 69.2 & 70.8 & 53.0 & 44.8         & 64.9        & 59.4                    & 72.7              \\
ASTER                   & ST+MJ                     & 600k                         & 93.6     & 89.0   & 93.0     & 74.2 & 80.0    & 85.4 & 69.7 & 71.0 & 53.1 & 47.1          & 65.4        & 60.2                    & 73.5          \\
ASTER                   & ST+MJ                     & 4m                        & \textbf{93.7}    & \textbf{89.5}   & \textbf{93.2}   & \textbf{74.7} & \textbf{81.1} & \textbf{86.1} & \textbf{69.8} & \textbf{71.6}  & \textbf{53.1} & \textbf{47.8}          & \textbf{65.5}        & \textbf{60.6}                    & \textbf{73.9}                \\
\hline
ABINet                   & ST+MJ                     & 100k                        & 95.7&94.5&97.2&85.7&88.4&90.3&76.4&81.1&63.1&62.2&76.1&66.1&81.4 \\
ABINet                   & ST+MJ                     & 600k                        & \textbf{96.1}&94.5&\textbf{97.8}&86.0&88.9&90.3&76.4&81.1&63.7&61.8&\textbf{76.9}&\textbf{67.6}&81.8 \\
ABINet                   & ST+MJ                     & 4m                        & 95.9     & \textbf{{94.9}}    & \textbf{{97.8}}   & \textbf{{86.1}}  & \textbf{{89.9}}  & \textbf{{91.4}} &    \textbf{{76.8}}   &    \textbf{{82.8}}   &    \textbf{64.0}   &  \textbf{{63.5}}      & 76.8      &           66.7          & \textbf{{82.3}}\\
\hline
\end{tabularx}}
\vspace{-1em}
\caption{Ablation study of the number of the audio data. Img. and Avg. are short for image and average, respectively. In each column, the best performance result of every method is shown in bold font.}
\label{tab:abl_audio_num}
\vspace{-1.5em}
\end{table*}

\begin{table*}[htbp!]
\centering
\small
\setlength\tabcolsep{1.9pt}
{
\begin{tabularx}{0.98\linewidth}{ccccccccccccccccc}
\hline
\multirow{2}{*}{Method} & \multirow{2}{*}{Label type} & \multirow{2}{*}{Img. data} & \multirow{2}{*}{Audio data} & \multicolumn{3}{c}{Regular} & \multicolumn{6}{c}{Irregular}        & \multicolumn{2}{c}{Occluded} & \multirow{2}{*}{WordArt} &\multirow{2}{*}{Avg.} \\ \cmidrule(lr){5-7} \cmidrule(lr){8-13} \cmidrule(lr){14-15} 
                        &                                        &                           &                             & IIIT5K    & SVT    & IC13   & IC15 & SVTP & CT80 & CTW & TT & COCO & HOST          & WOST         &                          \\ \hline
CRNN                    & Mel                                    & MJ                        & 600k                         & \textbf{82.0}    & \textbf{81.0}   & \textbf{88.3}  & \textbf{57.5} & \textbf{65.4} & \textbf{60.4} & \textbf{55.4} & \textbf{53.5} & \textbf{37.7} & \textbf{36.9}         & \textbf{53.0}        & \textbf{38.4}                    & \textbf{59.1}              \\
CRNN                    & Linear                                 & MJ                        & 600k                         & 80.7    & 80.1   & 87.0   & 56.3 & 62.2 & 56.6  & 54.2  & 51.8 & 37.6 & 32.6         & 48.6        & 36.1                    & 57.0              \\ \hline
ASTER                   & Mel                                    & ST+MJ                     & 600k                         & \textbf{93.6}     & \textbf{89.0}   & \textbf{93.0}     & \textbf{74.2} & \textbf{80.0}    & \textbf{85.4} & \textbf{69.7} & \textbf{71.0} & \textbf{53.1} & \textbf{47.1}          & \textbf{65.4}        & \textbf{60.2}                    & \textbf{73.5}          \\
ASTER                   & Linear                                 & ST+MJ                     & 600k                         & \textbf{93.6}    & 88.0   & 92.7  & 73.4 & 78.5 & 83.7 & 68.8 & 70.5 & 52.4 & 46.2         & 65.3        & 59.6                    & 72.7  \\
\hline
ABINet                   & Mel                                    & ST+MJ                     & 600k                         & \textbf{95.9}     & \textbf{94.9}   & \textbf{97.8}     & \textbf{86.1} & \textbf{89.9}    & \textbf{91.4} & \textbf{76.8} & \textbf{82.8} & \textbf{64} & \textbf{63.5}          & \textbf{76.8}        & \textbf{66.7}                    & \textbf{82.2}          \\
ABINet                   & Linear                                 & ST+MJ                     & 600k                         & 95.7    & 94.8   & 96.9  & 85.8 & 89.3 & 90.5 & 76.2 & 81.2 & 63.8 & 63.1         & 75.6        & \textbf{66.7}                    & 81.6  \\
\hline
\end{tabularx}}
\vspace{-1em}
\caption{Ablation study of the spectrogram format. Img. and Avg. are short for image and average, respectively. In each column, the best performance result of every method is shown in bold font.}
\label{tab:abl_mel}
\vspace{-1.5em}
\end{table*}

\subsection{Ablation Studies}

\noindent\textbf{The number of the audio decoder layers.} To study the relationship between the performance and the ability of speech information extraction, we compare the result of models which are implemented with a different number of audio decoder layers $N$. As shown in Table~\ref{tab:abl_layer}, we can see that $N=3$ gets the best performance, and the performance of $N=4$ decreases a lot compared to $N=3$.

\begin{figure}[htbp]
\centering 
\includegraphics[width=0.3\textwidth]{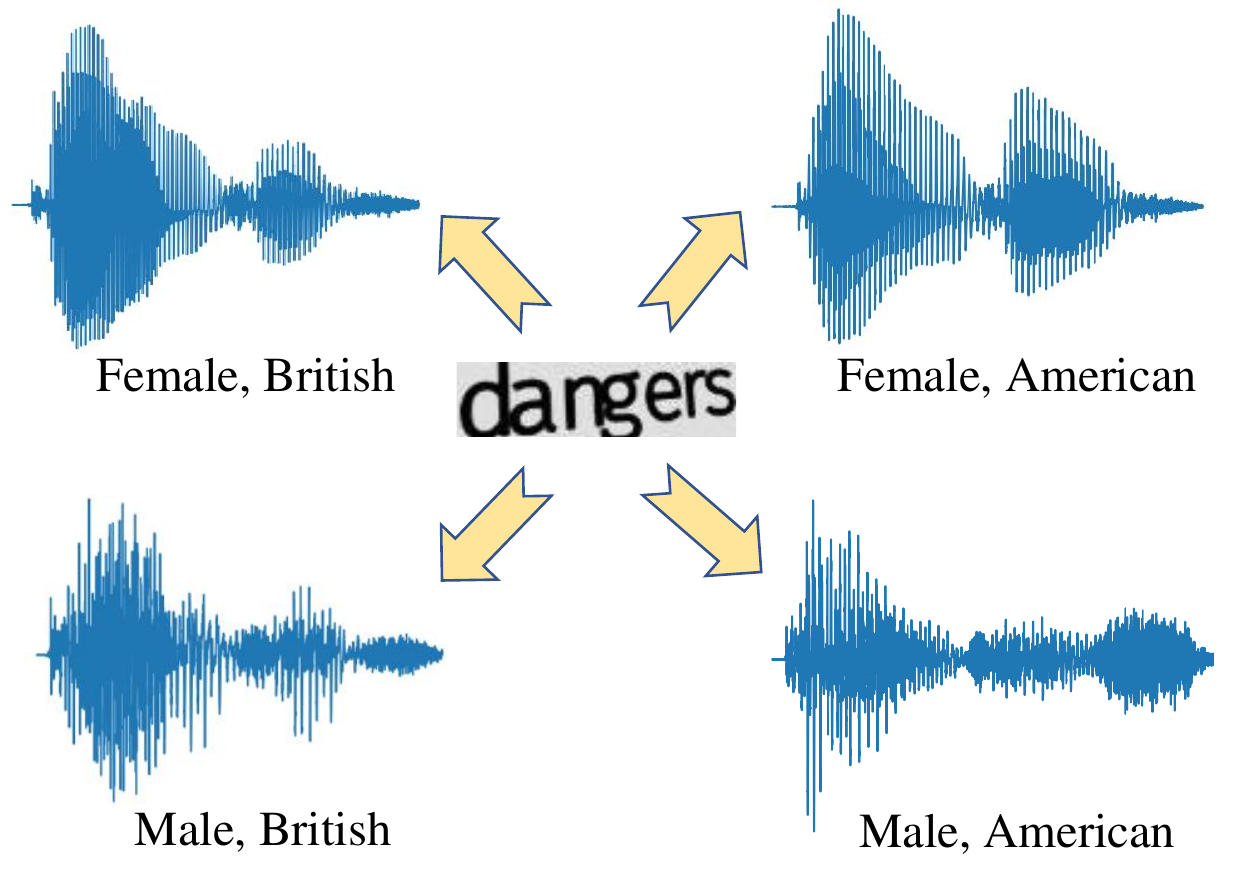}
\vspace{-1em}
\caption{The waveforms of different speakers. The female waveform is smoother and has a higher frequency than the male waveform.}
\vspace{-1.2em}
\label{fig:different_speaker_spec}
\end{figure}

\noindent\textbf{The influence of the accents.}
To examine the influence of the accents, we compare the performance of four different settings: American female accent, American male accent, British female accent, and British male accent. As shown in Table~\ref{tab:abl_type}, the guidance by the American female and British female audio achieves the best for CRNN, with 3.33\% higher than the baseline method in terms of the average accuracy. For ASTER, the best setting is using the British female audio, which is 1.9\% higher than the baseline ASTER. Note that for both methods and genders, the American and British accents bring similar gains for the baseline method, while the female accent is clearly more beneficial to the recognizers than the male accent.
By analyzing the sound, we find that the female voice is clearer while the male voice is deeper. Reflected in the waveforms, as shown in Figure~\ref{fig:different_speaker_spec}, the female waveform is smoother, clearly identifiable, and may easier to learn. In the following experiments, we adopt the British female audio.

\noindent\textbf{The influence of audio data size.}
To further demonstrate the effectiveness of audio information, we explore the influence of the audio data volume on text recognition.  As shown in Table~\ref{tab:abl_audio_num}, the increasing amount of data may gradually increase the recognition performance.

\noindent\textbf{The influence of spectrogram format.}
We perform a group of experiments on the spectrogram format including linear and mel scale to investigate how it affects the model performance. We report the results in Table~\ref{tab:abl_mel}, which shows that our method achieves the best performance under the mel spectrogram setting compared with linear spectrogram settings on all benchmarks. This is because the mel spectrogram has better discriminant validity for different audios than the linear spectrogram, and thus the audio decoder can easily capture the discrepancy of different text conditioned on the visual features. To this end, the audio decoder can facilitate better learning of discriminatory content by the image encoder.

\begin{figure}[htbp]
\centering
\includegraphics[width=0.4\textwidth]{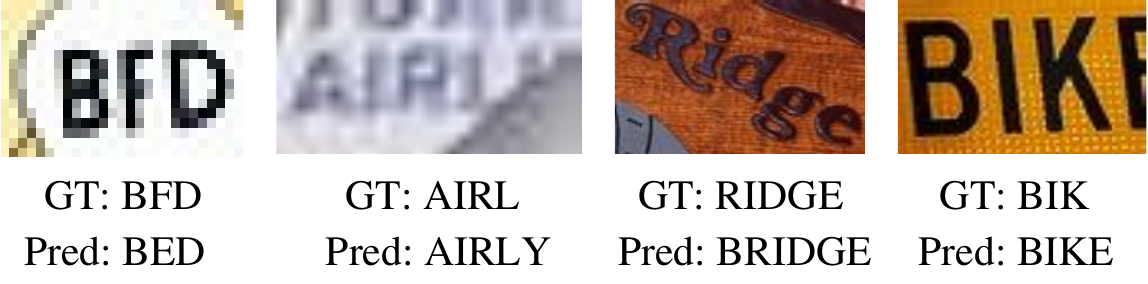}
\vspace{-1em}
\caption{Failure cases of our methods. GT and Pred. represent the ground-truth and the predicted text, respectively.}
\vspace{-1.1em}
\label{fig:failure_cases}
\end{figure}

\begin{figure}[htb]
\center
\resizebox{.9\columnwidth}{!}{
\begin{tabular}{cccc}
\hline
Input Images & Audio-guided MASTER & MASTER & GT \\
\hline

\raisebox{-0.5\height}{\includegraphics[width=0.24\linewidth,height=0.7cm]{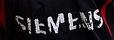}} 
& \makecell*[c]{  \textcolor[RGB]{21, 152, 56}{SIMENS}}
& \makecell*[c]{   \textcolor[RGB]{21, 152, 56}{S}\textcolor[RGB]{252, 54, 65}{H}\textcolor[RGB]{21, 152, 56}{E}\textcolor[RGB]{252, 54, 65}{I}\textcolor[RGB]{252, 54, 65}{V}\textcolor[RGB]{21, 152, 56}{ENS} } & SIMENS\\

\raisebox{-0.5\height}{\includegraphics[width=0.24\linewidth,height=0.7cm]{figs//predict_compare/984.png}} 
& \makecell*[c]{  \textcolor[RGB]{21, 152, 56}{Welcome}}
& \makecell*[c]{   \textcolor[RGB]{21, 152, 56}{Welc}\textcolor[RGB]{252, 54, 65}{ern}\textcolor[RGB]{21, 152, 56}{e} } & Welcome \\

\raisebox{-0.5\height}{\includegraphics[width=0.24\linewidth,height=0.7cm]{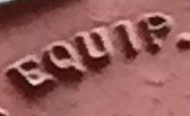}} 
& \makecell*[c]{  \textcolor[RGB]{21, 152, 56}{equip}}
& \makecell*[c]{  \textcolor[RGB]{252, 54, 65}{co}\textcolor[RGB]{21, 152, 56}{u}\textcolor[RGB]{252, 54, 65}{lf} } & equip\\

\raisebox{-0.5\height}{\includegraphics[width=0.24\linewidth,height=0.7cm]{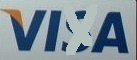}} 
& \makecell*[c]{  \textcolor[RGB]{21, 152, 56}{VISA}}
& \makecell*[c]{  \textcolor[RGB]{21, 152, 56}{VI}\textcolor[RGB]{252, 54, 65}{N}\textcolor[RGB]{21, 152, 56}{A} } & VISA\\

\hline

\end{tabular}
}
\vspace{-1em}
\caption{Recognition result samples. Green and red are correct and wrong predictions, respectively. More results are in appendix.}
\vspace{-1.8em}
\label{fig:predict_compare}
\end{figure}

\noindent\textbf{How useful on large-scale real data.}
As shown in Table~\ref{tab:real_data_textocr}, based on the large-scale real data TextOCR~\cite{singh2021textocr}, using 690k audio data of TextOCR can improve CRNN and ABINet from 56.0\% to 59.5\% and 78.2\% to 79.3\%, respectively, demonstrating the effectiveness when training on real data.

\noindent\textbf{Usefulness on sentence-level data vs. language modeling methods.} Despite the advantages of language modeling, our method provides further enhancement. We examined this using the line-level SROIE~\cite{huang2019icdar2019} dataset and ABINet that possesses language modeling capabilities. Incorporating audio into ABINet increased performance by 5.4\% over the original method (92.3\% vs. 86.9\%), demonstrating its effectiveness for text recognition when handling more sentences for actually text recognition.

\noindent\textbf{Usefulness vs. multiple decoders methods.} We assessed the performance of audio decoder against the use of ensemble decoders (CTC+Attention+Transformer) during the training stage. For a fair comparison, the parameters of these additional decoders were matched with those of the audio decoder. Subsequently, inference was carried out using one decoder with MASTER as the baseline. The average accuracy are 78.5\% and 77.7\%, respectively. This demonstrates the superiority of the proposed audio decoder over using multiple decoders.

\begin{table*}[htbp]
\centering
\small
\setlength\tabcolsep{2.2pt}
{
\begin{tabularx}{1.0\linewidth}{ccccccccccccccccc}
\hline
                         &                             &                              & \multicolumn{3}{c}{Regular}                                                                                           & \multicolumn{6}{c}{Irregular}                                                                                                                                                                                                                 & \multicolumn{2}{c}{Occluded}                                                  &                                       &                           &                       \\   \cmidrule(lr){4-6} \cmidrule(lr){7-12} \cmidrule(lr){13-14} 
\multirow{-2}{*}{Method} & \multirow{-2}{*}{Img. ddata} & \multirow{-2}{*}{Audio data} & IIIT5K                                & SVT                                   & IC13                                  & IC15                                  & SVTP                                  & CT80                                  & CTW                                   & TT                                    & COCO                                  & HOST                                  & WOST                                  & \multirow{-2}{*}{WordArt}             & \multirow{-2}{*}{TextOCR} & \multirow{-2}{*}{Avg.} \\ \hline 

CRNN                     & TextOCR                     & -                            &  77.4  & 79.3  & 88.1  & 58.1  & 62.2  & 61.8 & 54.9  & 52.6  & 41.7  & 23    & 41.3  & 36.3  & 51.5                      & 56.0           \\
CRNN                     & TextOCR                     & 690k                          & \textbf{78.9}                                  & \textbf{81.9}                                  & \textbf{90.4}                                  & \textbf{62.1}                                  & \textbf{66.4}                                  & \textbf{66.7}                                  & \textbf{56.8}                                  & \textbf{54.6}                                  & \textbf{43.4}                                  & \textbf{29.3}                                  & \textbf{48.8}                                  & \textbf{41.2}                                  & \textbf{53.2}                      & \textbf{59.5}           \\\hline
ABINet                   & TextOCR                     & -                            & 85.4                                  & 94             & 93.6           & 80.3           & \textbf{90.1}           & 90.6           & 71.3           & 79.7           & 63.7           & 60             & 72.6           & 66                                    & 70                        & 78.2           \\
ABINet                   & TextOCR                     & 690k                          & \textbf{86.03}                                 & \textbf{94.1}         & \textbf{95.4}           & \textbf{81.1}          & 89.8           & \textbf{92.1}           & \textbf{72.1}           & \textbf{80.5}           & \textbf{65.2}           & \textbf{61.26}         &\textbf{75.04}          & \textbf{68.2}                                  & \textbf{70.5}                      & \textbf{79.3}           \\ \hline
\end{tabularx}
}
\vspace{-1em}
\caption{Ablation study of training on real data. Img. and Avg. are short for image and average, respectively. In each column, the best performance result of every method is shown in bold font.}
\label{tab:real_data_textocr}
\vspace{-1.7em}
\end{table*}

\vspace{-0.8em}
\begin{figure}[htbp]
\centering
\includegraphics[width=0.4\textwidth]{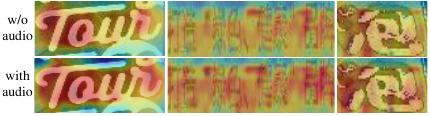}
\vspace{-1em}
\caption{Heat map of image features. Best view in screen.}
\vspace{-1.5em}
\label{fig:vis_feature_map}
\end{figure}

\noindent\textbf{Ablation study on non-English data.}
Our audio-guided text recognition method, primarily designed and evaluated for English scenarios, has been tested for universality with non-English data as well. Specifically, we selected datasets of images containing scene text in Latin German from MLT19~\cite{Nayef2019ICDAR2019RR}, and Chinese from ReCTS~\cite{Liu2019ICDAR2R}. These German and Chinese datasets comprise diverse pronunciations. We trained CRNN and ABINet on the German and Chinese datasets using the same procedure as for the English dataset. Table~\ref{tab:non_english_data} shows that our audio-guided method can improve the performance of existing methods by leveraging audio features for multilingual text recognition.

\begin{table}[htbp]
\centering
\small
\setlength\tabcolsep{2.2pt}
{
\begin{tabularx}{0.95\linewidth}{c|c|c|c}
\hline
Method & Audio data & MLT19-German & ReCTS-Chinese      \\ 
\hline
CRNN            & -          & 70.6                     & 63.1          \\
CRNN      & 600k        & 71.7                     & 64.8          \\
\hline
ABINet-LV       & -          & 74.8                     & 81.5          \\
ABINet-LV & 60k        & 75.6                     & 82.6   \\
\hline
\end{tabularx}
}
\vspace{-1em}
\caption{Ablation study on non-English data.}
\label{tab:non_english_data}
\vspace{-2.0em}
\end{table}

\noindent\textbf{Ablation study on out-of-vocabulary words.} Table~\ref{tab:oov_words} presents our investigation into the generalization capacity of the audio-guided method for out-of-vocabulary (OOV) data. We designed our OOV experimental setup to ensure that the words in the test set do not exist in the training data, thus facilitating an exploration of the method's generalization potential. Specifically, we selected French and German words from MLT19, composed solely of alphabetic characters, as our test set to guarantee they do not overlap with the English words in the training set. We utilized existing models trained on English datasets to directly test this set. The results demonstrated consistent improvements offered by AudioOCR. The conducted experiments provide empirical evidence supporting our method's capacity to generalize to OOV data. It's suggested that the model might learn pronunciation rules of vowels and consonants from the training data, enabling it to infer unseen words. Moreover, these findings suggest the feasibility of employing minimal audio data for real-world application training, thereby eliminating the need to cover all English words.

\vspace{-0.5em}
\begin{table}[htbp]
\centering
\small
\setlength\tabcolsep{1pt}
{
\begin{tabularx}{0.95\linewidth}{c|c|c}
\hline
Method            & French (3,883 imgs) & German (6,916 imgs) \\
\hline
CRNN                  & 70.9               & 76.1               \\
CRNN+Audio            & 72.2               & 76.5               \\
\hline
ASTER                & 77.8               & 79.9               \\
ASTER+Audio        & 78.3               & 82.2               \\
\hline
ABINet-LV            & 85.4               & 82.2               \\
ABINet-LV+Audio    & 86.1               & 82.5           \\
\hline
\end{tabularx}
}
\vspace{-1em}
\caption{Ablation study on out-of-vocabulary words.}
\label{tab:oov_words}
\vspace{-1.4em}
\end{table}

\noindent\textbf{Training cost of the proposed auxiliary task.}
For the training stage of baseline models, as shown in Table~\ref{tab:training_time}, using our method introduces an additional 19.4M, 0.44G, and 4.6 ms/image on average, respectively. It is valuable for real applications, since this auxiliary loss is only added during training and the training cost is acceptable, and no additional cost is incurred during the inference phase.

\vspace{-0.8em}
\begin{table}[htbp]
\centering
\small
\setlength\tabcolsep{2.2pt}
{
\begin{tabularx}{1.0\linewidth}{lcccc}
\hline
Model        & Params(M) & FlOPs(G) & ms/image & Acc. (\%)\\
\hline
MASTER       &     58.98      &    15.04      &     122.3   & 77.2       \\ %
MASTER+Audio &     78.42      &      15.47    &       125.8  & 78.5      \\ %
\hline
ABINet-LV       &      36.74     &       5.94   &      21.8  & 81.5       \\ %
ABINet-LV+Audio       &    56.17     &     6.39     &    27.5 & 82.3 \\  %
\hline
\end{tabularx}}
\vspace{-1em}
\caption{The training cost of the proposed auxiliary task.}
\label{tab:training_time}
\vspace{-1.2em}
\end{table}

\noindent\textbf{{Analyze the impact of common syllables.}} We conducted additional experiments with a dataset containing 45 common English prefixes and suffixes used in scene text collected from TT and CTW. We compared the performance of the ABINet method with and without audio. The results showed a 1.5\% performance improvement with the addition of audio. This indicates that the audio information, particularly the presence of commonly used syllables, contributes to the enhancement of recognition performance.

\noindent\textbf{{Qualitative Analysis.}}\label{sec:qual} As shown in Figure~\ref{fig:predict_compare}, compared to the baseline methods, audio-guided methods show more robust performance for complicated cases such as blurry or flipping images, distorted, occluded, curved, or art font text. We deduce this may be contributed by the spectrogram of corresponding text features in the frequency domain, which somewhat makes the visual representations more robust. We further visualize failure cases in Figure~\ref{fig:failure_cases}, from which we find the audio-guided recognizers tend to predict phonetically legal words even if the images missing partly characters. 
 
\noindent\textbf{How audio helps improve accuracy?} The use of audio information during training enhances the focus of the image encoder on the text areas, thereby boosting its ability to capture subtle visual signals. This is illustrated in Figure~\ref{fig:vis_feature_map}, where the heat map of the image features highlights larger text areas in both English and Chinese scenarios, even if Chinese has predominantly monosyllabic characters. This audio-visual fusion fortifies the discriminative capability of the image encoder, thereby improving text recognition accuracy. While not involved in inference, the encoder still uses training-derived information to extract relevant features. Hence, the audio decoder aids the recognition process by enriching the encoder with implicit text information.

\vspace{-0.5em}
\section{Conclusion}
In this work, we propose a simple yet effective plug-and-play module AudioOCR. To exploit the audio knowledge, we design a visual-audio decoder implemented by a transformer to excavate cross-modal information between visual and audio features for learning effective scene text visual representations. It can be easily connected to existing methods during the training phase meanwhile freely improve the performance of recognizers without extra cost in the inference stage. 
We evaluate the AudioOCR by combining it with 7 scene text recognition methods on 12 benchmarks including regular, irregular, and occluded datasets. The results demonstrate the consistent improvements contributed by the proposed AudioOCR. Besides, it can also deal with more challenging problems including non-English text, out-of-vocabulary words, and text with various accents.

Since the text-to-speech tool, PaddleSpeech, can only handle letters and does not support special symbols for now. The SynthAudio dataset only contains non-special symbolic audio. The improvement of AudioOCR may be further improved if such special symbols can also be guided by the audio, which leaves for future exploration. 

As the audio information is naturally connected to the scene text recognition information, the successful attempt of using the AudioOCR further sheds a broad space of cross-modal interaction for scene text recognition.

{\small
\bibliographystyle{ieee_fullname}
\bibliography{references}
}

\clearpage

\newpage
\appendix

\section{Appendix}

\subsection{Datasets}
\noindent\textbf{MJSynth} (MJ) is the synthetic text dataset proposed in~\cite{Jaderberg2014SyntheticDA} a.k.a. Syn90k. The dataset has 9 million images generated from a set of 90k common English words. Every image is annotated with word-level ground-truth. We randomly generate partial corresponding audio according to the text of the images.

\noindent\textbf{SynthText} (ST)~\cite{gupta2016synthetic} is a synthetic text dataset originally introduced for text detection. The generating procedure is similar to~\cite{Jaderberg2014SyntheticDA}, and the words are rendered onto a full image with a large resolution.
800 thousand full images are used as background images, and usually, each rendered image contains around 10 text lines. 
It is also widely used for scene text recognition via cropped images. We randomly generated partial corresponding audio according to the text of the images.

\noindent\textbf{SynAdd} (SA) is the synthetic text dataset proposed in~\cite{Li2019ShowAA}. The dataset contains 1.6 million word images using the synthetic engine proposed by~\cite{Jaderberg2014SyntheticDA} to compensate for the lack of special characters like punctuations.

\noindent\textbf{The Details of Benchmarks.} The detailed descriptions of the benchmarks we used are briefly shown in Table~\ref{tab:text_reco_benchmarks}.
\begin{table}[htbp]
\centering
\normalsize
\setlength\tabcolsep{1.8pt}
{
\begin{tabularx}{1.0\linewidth}{c|c|c|c}
\hline
ID & Benchmarks & Test images & Description      \\ 
\hline \#1 &\text { IIIT5K } & 3,000 & \text { regular scene text } \\
\hline \#2 &  \text { SVT } & 647 & \text { regular scene text } \\
\hline \#3 &\text { IC13 } & 1,015 & \text { regular scene text } \\
\hline \#4 &\text { IC15 } & 1,811 & \text { incidental scene text } \\
\hline \#5 &\text { SVTP } & 645 & \text { perspective scene text } \\
\hline \#6 &\text { CT80 } & 288 & \text { irregular scene text } \\
\hline \#7 &\text { CTW } & 1,572 & \text { irregular scene text } \\
\hline \#8 & \text { TT } & 2,201 & \text { irregular scene text } \\
\hline \#9 &\text { COCO } & 9,896 & \text { incidental scene text } \\
\hline \#10 &\text { WOST } & 2,416 & \text { weakly occluded scene text } \\
\hline \#11 &\text { HOST } & 2,416 & \text { heavily occluded scene text } \\
\hline \#12 &\text { WordArt } & 1,511 & \text { wordart scene text } \\
\hline
\end{tabularx}
}
\caption{Details of the scene text recognition benchmarks.}
\label{tab:text_reco_benchmarks}
\end{table}

\noindent\textbf{The Details of SynthAudio.} The detailed information of the SynthAudio can be found in Table~\ref{tab:audio_data}.

\begin{table}[htbp]
\centering
\begin{tabular}{c|c|c|c}
\hline
ID & Text Source & Audio Num & Audio Type       \\ \hline
\#1 & MJ          & 600,000   & Female, British  \\ \hline
\#2 & MJ          & 600,000   & Female, American \\ \hline
\#3 & MJ          & 600,000   & Male, British    \\ \hline
\#4 & MJ          & 600,000   & Male, American   \\ \hline
\#5 & MJ          & 4,000,000 & Female, British  \\ \hline
\#6 & ST          & 4,000,000 & Female, British  \\ \hline
\#7 & COCO   & 42,142    & Female, British  \\ \hline
\#8 & IC11        & 3,567     & Female, British  \\ \hline
\#9 & IC13        & 848       & Female, British  \\ \hline
\#10 & IC15        & 4,468     & Female, British  \\ \hline
\#11 & IIIT5K      & 2,000     & Female, British  \\ \hline
\#12 & TextOCR      & 690,000     & Female, British  \\ \hline
\#13 & MLT19      & 600,000      & Female, German  \\ \hline
\#14 & ReCTS      & 600,000      & Female, Chinese  \\ \hline
\end{tabular}
\caption{Details of the generated SynthAudio data. British and American represent two different accents. }
\label{tab:audio_data}
\end{table}

\subsection{Implementation Details}
\noindent\textbf{The Details of English and Chinese Audio Generation.} We use PaddleSpeech~\cite{zhang2022paddlespeech}\footnote{\url{https://github.com/PaddlePaddle/PaddleSpeech}} engine to convert the English and Chinese text labels to the audio file via neural network based text to speech (TTS) models. The procedure of TTS models first generates mel spectrograms from text sequences using an acoustic model, and then synthesizes speech from the generated mel spectrograms using a separately trained vocoder. The configurations for audio generation are shown in Table~\ref{tab:audio_config}. Specifically, we chose Fastspeech2~\cite{Ren2021FastSpeech2F} as our acoustic model, which can remarkably promote the speed of generating audio data. The acoustic model can represent the relationship between an audio signal and the phonemes or other linguistic units that make up speech. The vocoder is used to convert mel spectrograms to an audio waveform. We chose HiFiGAN~\cite{Kong2020HiFiGANGA} to be our vocoder since it is commonly used in academic and industrial scenarios. The text corpus is in English and directly from existing synthesized datasets like MJ~\cite{Jaderberg2014SyntheticDA}, ST~\cite{gupta2016synthetic}, and real data including COCO-Text, IC11, IC13, IC15, IIIT5K, TextOCR, and ReCTS-Chinese~\cite{Liu2019ICDAR2R}, which are commonly served as training data for scene text recognition. The generated audio will be used to generate the mel spectrogram $\mathbf{Z}$.

\noindent\textbf{The Details of German Audio Generation.} Because PaddleSpeech does not support German audio generation, we use Google Text-to-Speech API engine to convert the German text label to the audio file including MLT19-German~\cite{Nayef2019ICDAR2019RR}. The configuration of the Google Text-to-Speech API is depicted in Table~\ref{tab:google_tts_api}.

\noindent\textbf{The Details of Mel Spectrogram Generation.} The generation configurations of the mel spectrogram $\mathbf{Z}$ are shown in Table~\ref{tab:spec_config}. The sample ratio reflects the frequency we sample from the original audio signal, a higher sample ratio tends to deliver a better-quality audio reproduction. Here we set it to 22050, which means we sample 22050 times in one second. For the window function in Eq.~\ref{eq:stft1}, we set the window stride to 275.625 (ms) and the window length $W$ to 1102.5 (ms), which decides how many times we compute the Fast Fourier Transform (FFT) on the origin audio signal, and the dimension of FFT is set to 2048. The dimension of the mel spectrogram $F$ is set to 80. The higher the dimension of the mel spectrogram is, the more information can be learned from the audio data, but it also brings extra computation costs.
\begin{equation}
\mathtt{STFT}\{\bm{x}\}(f, l)=\sum_{t=0}^{T}\bm{x}_t\cdot \bm{w}_{t-l}\cdot \mathtt{exp}\left(-i \frac{2 \pi t}{T} f \right) \,.
 \label{eq:stft1}
\end{equation}

\begin{table}[htbp]
\centering
\begin{tabular}{c|c}
\hline
Parameters        &      Configuration       \\ \hline
Acoustic Model    & Fastspeech2 \\
Vocoder           & HiFiGAN     \\
Text Corpus           & MJ+ST+Real    \\
Language & English/Chinese \\ \hline
\end{tabular}
\caption{The configurations of PaddleSpeech for English/Chinese audio generation. Real consists of COCO, IC11, IC13, IC15, IIIT5K, TextOCR, and ReCTS-Chinese~\cite{Liu2019ICDAR2R}.}
\label{tab:audio_config}
\vspace{-0.3cm}
\end{table}

\begin{table}[htbp]
\centering
\begin{tabular}{c|c}
\hline
Parameters        &      Configuration       \\ \hline
Voice Type    & Neural2 \\
Language Code           & de-DE     \\
Voice Name           & de-DE-Neural2-F     \\
SSML Gender          & FEMALE     \\
Text Corpus           & MLT19-German    \\
Language & German \\ \hline
\end{tabular}
\caption{The configurations of Google Text-to-Speech API for German audio generation.}
\label{tab:google_tts_api}
\vspace{-0.3cm}
\end{table}

\begin{table}[htbp]
\centering
\begin{tabular}{c|c}
\hline
Parameters        &     Configuration        \\ \hline
Sample Ratio (SR) & 22050       \\
Window Stride     & 275.625 (ms)     \\
Window Length $W$    & 1102.5 (ms)   \\
Dimension of FFT        & 2048  \\
Dimension of Mel Spectrogram   $F$    & 80   \\
  \hline
\end{tabular}
\caption{The generation configurations of the mel spectrogram $\mathbf{Z}$.}
\label{tab:spec_config}
\vspace{-0.25cm}
\end{table}

\subsection{More Results.}

\noindent\textbf{Ablation study for the weight of losses.}
As shown in Table~\ref{tab:abl_loss_weight}, we vary the trade-off audio loss weight $\alpha$ ranging in [0.5, 1, 2], observing that $\alpha=1$ can get the best performance.
We guess that $\alpha=2$ may break the balance between visual information and audio information, and $\alpha=0.5$ results in insufficient training using the audio feature. %

\begin{figure}[htb]
\center
\resizebox{.99\columnwidth}{!}{
\begin{tabular}{cccc}
\hline
Input Images & Audio-guided MASTER & MASTER & GT \\
\hline

\raisebox{-0.5\height}{\includegraphics[width=0.24\linewidth,height=0.7cm]{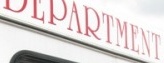}} 
& \makecell*[c]{  \textcolor[RGB]{21, 152, 56}{DEPARTMENT}}
& \makecell*[c]{   \textcolor[RGB]{252, 54, 65}{S}\textcolor[RGB]{21, 152, 56}{ARTMENT}} & DEPARTMENT\\

\raisebox{-0.5\height}{\includegraphics[width=0.24\linewidth,height=0.7cm]{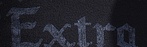}} 
& \makecell*[c]{  \textcolor[RGB]{21, 152, 56}{EXTRA}}
& \makecell*[c]{   \textcolor[RGB]{21, 152, 56}{EXTR}\textcolor[RGB]{252, 54, 65}{N}  } & EXTRA\\

\raisebox{-0.5\height}{\includegraphics[width=0.24\linewidth,height=0.7cm]{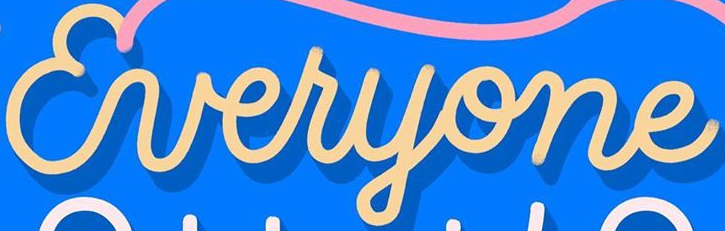}} 
& \makecell*[c]{  \textcolor[RGB]{21, 152, 56}{everyone}}
& \makecell*[c]{  \textcolor[RGB]{21, 152, 56}{e}\textcolor[RGB]{252, 54, 65}{uc}\textcolor[RGB]{21, 152, 56}{yone} } & everyone\\

\raisebox{-0.5\height}{\includegraphics[width=0.24\linewidth,height=0.7cm]{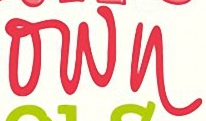}} 
& \makecell*[c]{  \textcolor[RGB]{21, 152, 56}{own}}
& \makecell*[c]{   \textcolor[RGB]{21, 152, 56}{ow}\textcolor[RGB]{252, 54, 65}{y}  } & own\\

\raisebox{-0.5\height}{\includegraphics[width=0.24\linewidth,height=0.7cm]{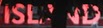}} 
& \makecell*[c]{  \textcolor[RGB]{21, 152, 56}{island}}
& \makecell*[c]{  \textcolor[RGB]{21, 152, 56}{is}\textcolor[RGB]{252, 54, 65}{en}\textcolor[RGB]{21, 152, 56}{nd}\textcolor[RGB]{252, 54, 65}{d} } & island\\

\raisebox{-0.5\height}{\includegraphics[width=0.24\linewidth,height=0.7cm]{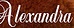}} 
& \makecell*[c]{  \textcolor[RGB]{21, 152, 56}{alexandra}}
& \makecell*[c]{  \textcolor[RGB]{21, 152, 56}{alexand}\textcolor[RGB]{252, 54, 65}{i}\textcolor[RGB]{21, 152, 56}{a} } & alexandra\\

\raisebox{-0.5\height}{\includegraphics[width=0.24\linewidth,height=0.7cm]{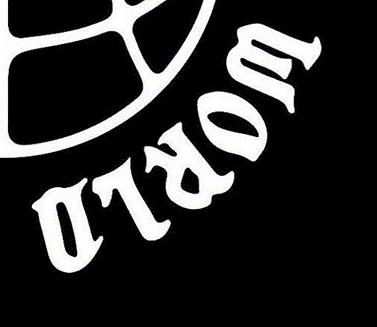}}
& \makecell*[c]{  \textcolor[RGB]{21, 152, 56}{WORLD}}
& \makecell*[c]{  \textcolor[RGB]{252, 54, 65}{140.1221.21} } & WORLD\\
\hline

\end{tabular}
}
\vspace{-1em}
\caption{Samples of recognition results of audio-guided and original methods. Green and red characters represent correct and wrong predictions, respectively. }
\vspace{-1.5em}
\label{fig:predict_compare_app}
\end{figure}

\noindent\textbf{Qualitative Analysis.} Fig.~\ref{fig:predict_compare_app} are more qualitative samples of recognition results of audio-guided and original methods. 

\begin{table*}[htbp]
\centering
\small
\setlength\tabcolsep{2.0pt}
{
\begin{tabularx}{1.0\linewidth}{ccccccccccccccccc}
\hline
\multirow{2}{*}{Method} & \multirow{2}{*}{Loss weight} & \multirow{2}{*}{Img. data} & \multirow{2}{*}{Audio data} & \multicolumn{3}{c}{Regular} & \multicolumn{6}{c}{Irregular}        & \multicolumn{2}{c}{Occluded} & \multirow{2}{*}{WordArt} & \multirow{2}{*}{Avg.} \\ \cmidrule(lr){5-7} \cmidrule(lr){8-13} \cmidrule(lr){14-15} 
                        &                                    &                           &                             & IIIT5K    & SVT    & IC13   & IC15 & SVTP & CT80 & CTW & TT & COCO & HOST          & WOST         &                          \\ \hline
CRNN                    & 1                                & MJ                        & 600k                         & \textbf{82.0}    & \textbf{81.0}   & \textbf{88.3}  & \textbf{57.5} & \textbf{65.4} & \textbf{60.4} & 55.4 & \textbf{53.5} & \textbf{37.7} & \textbf{36.9}         & \textbf{53.0}        & \textbf{38.4}                    & \textbf{59.1}              \\
CRNN                    & 2                                & MJ                        & 600k                         & 81.2    & 80.2   & 87.2  & 56.7 & 62.2 & 58.7 & 54.8 & 52.1 & 37.5 & 33.6         & 48.5        & 35.4                    & 57.3          \\
CRNN                    & 0.5                              & MJ                        & 600k                         & 81.3     & 80.6   & 87.0     & 57.3 & 62.3 & 60.1 & \textbf{55.6}  & 52.5 & 37.3 & 34.3         & 49.2        & 37.7                    & 57.9              \\ \hline
ABINet                   & 1                                & ST+MJ                     & 4m                         & \textbf{95.9}     & \textbf{94.9}   & \textbf{97.8}     & \textbf{86.1} & \textbf{89.9}    & \textbf{91.4} & \textbf{76.8} & \textbf{82.8} & \textbf{64} & \textbf{63.5}          & \textbf{76.8}        & \textbf{66.7}                    & \textbf{82.2}          \\
ABINet                   & 2                                & ST+MJ                     & 4m                         & 95.8    & \textbf{94.9}   & 96  & 85.8  &  89.6  & 90.3  & 76.9  & 81.6  & 63.4  & 63.1          & \textbf{76.8}        & \textbf{66.7}                    & 81.7              \\
ABINet                   & 0.5                              & ST+MJ                     & 4m                         & 95.8    & 94.7   & 96.9  & 85.3  & 89.8  & 90.2  & 76.7  &  81.4  & 63.9  & 62.9        & 76.5        & 66.5                    & 81.7   \\
\hline
\end{tabularx}}
\vspace{-0.5em}
\caption{Ablation study of loss weight of audio decoder. Img. and Avg. are short for image and average, respectively. In each column, the best performance result of every method is shown in bold font.}
\label{tab:abl_loss_weight}
\vspace{-1em}
\end{table*}

\end{document}